# ROMANIAN SPEECH RECOGNITION EXPERIMENTS FROM THE ROBIN PROJECT

ANDREI-MARIUS AVRAM, VASILE PĂIȘ, DAN TUFIȘ

*Research Institute for Artificial Intelligence, Romanian Academy*
*{andrei.avram,vasile,tufis}@racai.ro*

## Abstract

One of the fundamental functionalities for accepting a socially assistive robot is its communication capabilities with other agents in the environment. In the context of the ROBIN project, situational dialogue through voice interaction with a robot was investigated. This paper presents different speech recognition experiments with deep neural networks focusing on producing fast (under 100ms latency from the network itself), while still reliable models. Even though one of the key desired characteristics is low latency, the final deep neural network model achieves state of the art results for recognizing Romanian language, obtaining a 9.91% word error rate (WER), when combined with a language model, thus improving over the previous results while offering at the same time an improved runtime performance. Additionally, we explore two modules for correcting the ASR output (hyphen and capitalization restoration and unknown words correction), targeting the ROBIN project's goals (dialogue in closed micro-worlds). We design a modular architecture based on APIs allowing an integration engine (either in the robot or external) to chain together the available modules as needed. Finally, we test the proposed design by integrating it in the RELATE platform and making the ASR service available to web users by either uploading a file or recording new speech.

*Key words* — automatic speech recognition, Romanian, natural language processing, deep neural networks.

## *1. Introduction*

ROBIN[1] is a user-centred project designing software systems and services for the use of robots in an interconnected digital society. The project covers a diverse range of robots: assistive robots for the support of people with special needs, social robots for interaction with store customers, and software robots that can be installed on intelligent vehicles to achieve autonomous car driving.

One of the subprojects, called ROBIN-Dialog[2] aims to develop a series of scenarios for several micro-worlds, as well as to develop the technology of processing the Romanian language for situational dialogues in these micro-worlds. In this context, voice interaction with the dialog system of the robot becomes very important. The end-result involves linking together an automatic speech recognition (ASR) system, a dialog management system and a text to speech (TTS) system. This allows a user to interact with the robot using only spoken language.

---

[1] http://aimas.cs.pub.ro/robin/en/

[2] http://aimas.cs.pub.ro/robin/en/robin-dialog/

Because the end system has several components (ASR, dialog component, TTS), each component must exhibit low latency and execute in near real-time to improve the experience of the user. This paper focuses on ASR system experiments performed with the goal of achieving a latency as low as possible while still obtaining state-of-the-art (SOTA) results. The module itself was developed and tested outside of the dialog system, while providing application programming interfaces (API) that allows it to be integrated into other complex systems.

This paper is structured as follows: we start in Section 2 by presenting related work, including previous work within the ROBIN project, followed by Section 3 with the proposed modular system architecture. Then, in Section 4 we describe the datasets used, followed in Sections 5 and 6 by a description of the ASR system and the implemented modules. Section 7 presents the evaluation results. Finally, the conclusions and possible future work directions are presented in Section 8.

## 2. *Related work*

Automatic Speech Recognition (ASR) consists in translating human spoken utterances into a textual transcript, and it is a key component in voice assistants (Lopatovska *et al.*, 2019), in spoken language translation systems (Di Gangi *et al.*, 2019) or in generating automatic transcriptions for audio and videos (Noda *et al.*, 2019). Most of the ASR systems before the deep learning revolution used variations of Hidden Markov Models (HMM) (Garg *et al.*, 2016), and although they achieved good Word Error Rates (WER), they became very slow for large vocabularies and could not be used for open domain real-time transcriptions. ASR systems can be classified into end-to-end and pipeline systems. End-to-end systems can get as input both raw audio wave and handcrafted features, while pipeline systems have specific components (assembled into a processing pipeline) to extract speech features. Both system types can benefit from additional text correction modules.

Georgescu et al. (2019) considered the application of neural networks to Romanian ASR systems using the Kaldi[3] toolkit. As described in the paper, the authors evaluated the model on two corpora: RSC-eval and SSC-eval, achieving WERs of 2.79% and 16.63%, respectively. Even though the two evaluation sets represent different types of speech (read speech and spontaneous speech), for our implementation we are interested in a general ASR system, working regardless of the speech type. Hence, for comparison purposes we consider the average WER of both evaluation corpora, thus leading to an average WER of 9.71%. Furthermore, the Kaldi toolkit uses a pipeline system, where each component is treated as an independent module, so, with regards to latency, this approach has the disadvantage of each module adding its own latency to the overall process.

For the purposes of the ROBIN project, a similar approach based on the Kaldi toolkit was considered in Tufiș et al. (2019a). Preliminary results, reported in the paper, were rather modest, yielding a WER of approximately 25%, which is larger than the WER reported by Georgescu et al. (2019) on the SSC-eval corpus. Nevertheless, this can be attributed to the different audio corpus used for training the model, rather than to the technology itself.

---
[3] https://kaldi-asr.org

One of the recent end-to-end speech recognition architectures is DeepSpeech2 (Amodei *et al.*, 2016). According to the research paper, models trained with this toolkit were able to achieve a WER of less than 10% for both English and Mandarin on several evaluation datasets. To achieve this result, 11.940 hours of English speech data was fed through a deep neural network consisting of 11 layers, while for Mandarin the training dataset consisted of 9.400 hours. Additionally, in a previous research (Avram et al., 2020), we also confirmed that this architecture is capable of obtaining a WER less than 10% for the Romanian language, while considering both read and spontaneous speech.

Comparing with the large volume of speech data used by Amodei et al. (2016), totalling around 10.000 hours for each of the investigated languages, data available within the ROBIN project for the Romanian language is far smaller, totalling around 230 hours of audio aligned with text. Nevertheless, we considered experimenting with only this amount of data, trying to construct an end-to-end ASR system for Romanian.

## 3. *System architecture*

Given the goal of integrating the resulting ASR system into the larger ROBIN-Dialog context, the implementation needs to encapsulate the functionality into a dedicated module and expose it via easy to use APIs. Furthermore, taking into account that envisaged human-robot dialogues are well defined, based on a closed-world scenario, controlled by a Dialog Manager (DM) (Ion et al., 2020), the ASR system can be complemented by different correction models, further improving the recognized speech. These correction models are also exposed as APIs and invoked as needed.

For API invocation we considered using the HyperText Transfer Protocol (HTTP) GET and POST requests. Furthermore, the implementation adheres to a stateless, client-server, Representational State Transfer (REST) design. Finally, the result of an API call will be provided in JSON format. Thus, invoking a method involves sending a standard HTTP request with the required parameters sent according to the specific method (GET or POST) and parsing the output JSON to extract the results. Since high-performance JSON parsing libraries exist for most of the programming languages, this encoding will not add significant latency to the overall process.

The following modules with their corresponding methods and parameters were envisaged:

a) ASR: "/transcribe" method uses a single parameter: "file". The method receives a WAV file in the parameter via the POST HTTP request, and then it invokes the ASR model to transcribe the received file. Finally, it returns a JSON with two fields: "status" representing the success or error of the call and the corresponding transcription in the "transcription" field.

b) Hyphen restoration and capitalisation: "/correct" method receives a "text" parameter via either HTTP GET or POST. The received text is supposed to be the result of the ASR module "/transcribe" method described above. The method checks for missing hyphens and tries to restore them. The output JSON contains a "status" field representing the success or error of the call, a "text" field with the

    corrected text and an optional "comments" field used for analysing the correction process (not needed for integrations).

c) Additional language model-based correction for unknown words: "/correct" method behaving similarly to the previous method, receiving a "text" parameter and returning an output JSON with "status", "text" and optional "comments".

We opted for an architecture with low cohesion, so we isolated each module and created a chain of API calls that are invoked by the integration system, as depicted in Figure 1. This design was considered because it also allowed the integration system to select the calls and their order depending on the context. For example, it can be considered that very short transcriptions, consisting of only one or two words may not need hyphen and capitalization correction, while they may still employ other corrections. Additionally, the design allows evaluation and timing of each individual module.

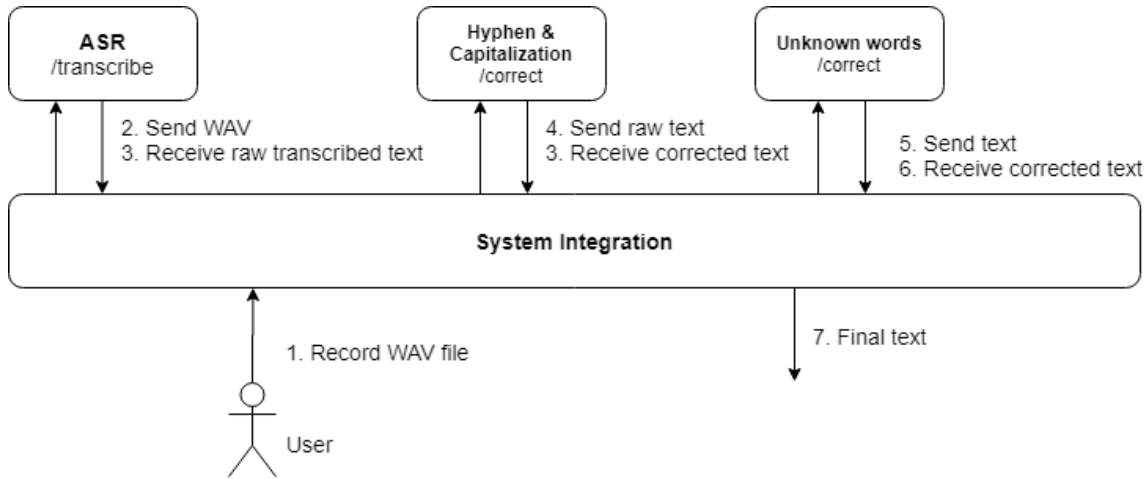

**Figure 1.** System integration architecture

## 4. Datasets

Two kinds of corpora were needed for the purposes of this work: first, a multimodal corpus containing high quality alignment of speech to text that was used to train the actual ASR system; second, large text corpora that were used to train the different language models for implementing additional corrections.

The main audio resource used was the speech component of the representative corpus of contemporary Romanian language (CoRoLa) (Tufiș et al., 2019b). CoRoLa has been jointly developed, as a priority project of the Romanian Academy, by two institutions: "Mihai Drăgănescu" Research Institute for Artificial Intelligence (from Bucharest) and the Institute of Computer Science (from Iași). The oral texts in CoRoLa are mainly professional recordings from various sources (radio stations, recording studios). They are accompanied by the written counterpart: the transcription either from their provider or made by the project partners. Therefore, different principles applied in their transcription, potentially making it more difficult to use for the purposes of the ASR system. Another part of the oral corpus is represented by read texts: read news in radio stations, texts read by professional speakers recorded in studios, and extracts from Romanian Wikipedia read by non-professionals, by volunteers, recorded in non-professional environments. In their case, the written component is provided by the sources, or was collected by the project partners (Mititelu et al., 2018).

The speech component of the CoRoLa corpus can be interrogated by means of the Oral Corpus Query Platform (OCQP)[4]. This allows searching for words and listen to their spoken variant, based on the alignment between text and speech (Tufiș et al., 2019b).

In the context of the RETEROM project[5], the CoBiLiRo platform (Cristea et al., 2020) was built to allow gathering of additional bimodal corpora with one of the final goals being to enrich the CoRoLa corpus. Thus, additional corpora with speech and text alignments were employed. This includes: Romanian Digits (RoDigits) (Georgescu *et al.*, 2018), Romanian Speech Synthesis (RSS) (Stan et al., 2011), Romanian Read Speech Corpus (RSC) (Georgescu et al., 2020).

Additionally, the Common Voice corpus (Ardila *et al.*, 2019) was considered, being a massively multilingual dataset of transcribed speech that, as of October 2020, contains over 7,300 hours of validated audio in 54 languages from over 50,000 speakers. The Romanian version is one of the recently added languages and its corresponding corpus contains 7 hours of transcribed audio recorded by 79 speakers, from which only 5 hours are validated. The corpus sentences were collected from Wikipedia using a sentence collector, and each sentence must be approved by two out of three reviewers before reaching the final version of the corpus, thus allowing for an acceptable audio quality aligned with the short text representing the sentence.

The main text resource used for language models was represented by the CoRoLa corpus. It is a large, growing, collection of Romanian texts, currently containing 941,204,169 tokens. Currently, all texts are more recent than 1945, therefore CoRoLa is a contemporary corpus. Various annotation levels were employed and the corpus can be queried through various interfaces (Cristea et al., 2019), including KorAP (Banski et al., 2012).

In addition to the texts from CoRoLa, we considered the OSCAR corpus (Suárez et al., 2019). It is an huge open-source multilingual corpus that was obtained by filtering the Common Crawl[6] and by grouping the resulting text by language. The Romanian version contains approximately 11 GB of deduplicated shuffled sentences. Even though CoRoLa is a representative corpus of the Romanian language, we considered that adding more text to the training of a language model could benefit in terms of accuracy.

## 5. *ASR system*

Following the DeepSpeech2 architecture, we first computed the Mel-frequency cepstral coefficients (MFCC) (Logan et al., 2000) on fixed-size audio windows of 20 ms. Then, the resulted spectrogram was fed into a deep neural network model. However, since the model presented by (Amodei *et al.*, 2016) made use of a large number of speech hours we considered a simpler model, consisting of only 8 neural layers: 2 convolutional 2D layers, 4 layers with bidirectional long short-term memory (BiLSTM) (Hochreiter et al., 1997), 1 lookahead convolution and a final fully connected (FC) layer. Batch normalization (BN) (Ioffe et al., 2015) was then used after each layer except the last one for faster model convergence and more stability during training. Also, we used the Hard Hyperbolic Function (HardTanh) activation function before each (BN), a cheaper and more computationally efficient version of tanh, as proposed in (Xiang et al, 2017).

---

[4] http://corolaws.racai.ro/corola_sound_search/index.php

[5] http://www.racai.ro/p/reterom/

[6] https://commoncrawl.org/

The output of the network is a vector of size 33 that is computed using the FC layer and the softmax activation and it represents a distribution of probabilities over the 31 Romanian characters, together with the space character and the blank index. The "blank" index is a special character used for cases where a certain character is repeated. For example, "acceptat" is encoded as "ac_ceptat" to mark not to collapse the character "c" while decoding. Furthermore, to account for situations where the utterance of a character may take more than the size of a window, the Connectionist Temporal Classification (CTC) (Graves et al., 2006) loss was used to train the network. Additionally, the Adaptive Moment Estimation (Adam) optimizer (Kingma et al., 2014) was used, initially with a high learning rate to accelerate the training, and then applying a learning rate decay of 5% after each epoch to avoid oscillations in the later stages of training (You et al., 2019).

Following the paper of Amodei et al. (2016), for the English language, we incorporated a 5-gram language model for transcription correction in the ASR system, that was trained using the KenLM toolkit[7]. The toolkit allows the configuration of two hyperparameters: alpha (α)- that controls the contribution of the model to predicting a word, and beta (β) - the probability of inserting a new word into the sequence. Thus, during inference we search for the most probable transcription using (1):

$$Q(y) = \log(p_{ctc}(y|x)) + \alpha \log(p_{lm}(y)) + \beta word\_count(y) \qquad (1)$$

where $x$ is the audio input, $y$ is the output of the neural model, $p_{ctc}$ is the probability given by the CTC decoder, $p_{lm}$ is the probability given by the language model, $word\_count$ is the number of words in the predicted transcription, and α and β are the hyperparamters of the language model.

For the purposes of building this language model, the available text resources were first pruned to remove potentially wrong data, especially with regards to the OSCAR corpus. This involved applying removal rules such as: removing very short (less than 20 characters) or very long lines, lines with words containing no diacritics, lines with URLs. Additionally, known abbreviations and measurement units were replaced with their complete textual representation. This resulted in about 10 GB of "cleaned" raw text from the combination of CoRoLa and OSCAR.

The resulting system was transformed into a REST API server using the Flask framework[8] and the Waitress web server[9], according to the API specification described in Section 3. The service is making use of a configuration file, allowing tuning of the system parameters such as the beam width, whether to use or not the language model, or whether to use a GPU or only run on the CPU. The expected "file" parameter must represent a WAV file with the following characteristics: mono, 16-bit, 16 KHz.

---

[7] https://github.com/kpu/kenlm

[8] https://flask.palletsprojects.com/en/1.1.x/

[9] https://docs.pylonsproject.org/projects/waitress/en/stable/

## 6. *Additional text corrections*

As described in Section 3, several text correction mechanisms for the ASR output were envisaged. These aim to improve the results of the speech recognition system. They are based on n-gram models, trained on the CoRoLa corpus.

Since our neural network model generates characters (and not complete words), this may include situations were a hyphen is normally used but it is not present in the output. In some cases, such as "ți-am" vs "țiam" the correction is easy to perform, since "țiam" is not a valid Romanian word. However, in cases such as "s-a" vs "sa" the correction is more difficult since both words are valid in Romanian. Therefore, a more complex approach, based on context, is needed. For this purpose, we employed first a bigram model considering the frequency of using the current word ($W_k$) with and without hyphen together with the next word ($W_{k+1}$). If for some reason the bigram ($W_k$, $W_{k+1}$) is not available (possibly due to the following word being recognized incorrectly) we fall back to a unigram model, replacing the current word ($W_k$) with the most frequent form.

Basic capitalization restoration is performed using name lists. We considered reduced name lists containing mostly people names and locations (countries, large cities) to reduce the risk of capitalizing a word simply because it looks like a named entity. As opposed to named entity recognition, in this case we do not need to actually identify in text the corresponding entity type, since regardless of the type person names and location names will be written with first letter capitalized.

Since other recognition errors, apart from hyphens, may happen, we further considered an additional correction model making use of the Levenshtein distance to correct unknown words. For this purpose, we consider as unknown the words not appearing in the vocabulary generated from the CoRoLa corpus, considering a minimum number of occurrences of 10. When an unknown word Wk is encountered, the system will look for a bigram ($W'_k$, $W_{k+1}$) or ($W_{k-1}$, $W'_k$) having the Levenshtein distance between $W_k$ and $W'_k$ less than a certain threshold. If no bigram is identified, we fall back to a unigram model, checking the Levenshtein distance with other words in the vocabulary, having similar size (less than the considered threshold).

The text corrections were implemented in two modules: one for hyphen restauration and capitalization and one for the unknown words. Both were implemented in Java and exposed as REST APIs according to the description given in Section 3.

## 7. *Results*

For system evaluation we used a server with a Xeon 4210 CPU running at 2.2 Ghz and a single Quadro RTX 5000 GPU. As presented in the Introduction, we were interested in obtaining a model with low runtime latency. The ASR system's average recognition time on audio samples of less than 25 seconds was 600 ms, when running on the CPU, and 70 ms, when running on the GPU. With regard to space requirements, the DeepSpeech2 model file size is 160 MB and the uncompressed KenLM file size is 10.6 GB, both models occupying around 5.9 GB in RAM.

From the overall speech corpus, we extracted 5,000 samples that were used as the test set and another 5,000 samples that were used as the development set. All the audio files in both sets were less than 25 seconds in length, similar to the maximum expected ROBIN interactions.

For the purposes of the first language model employed in combination with the raw ASR, we used a grid search on the development dataset to find the best values, for α in the [0, 1.5] interval and for β in the [0, 3] interval with discrete steps of 0.1. The results of the grid search are depicted as a surface plot in Figure 2. As it can be observed, α has the highest influence on the overall WER, while β acts just as a regularizer for the predicted transcription, having a lower influence. The optimum for the hyperparameters is represented by the values of 0.3 for α and 1.5 for β. Using the combined ASR and language model (with the optimal parameters) the overall result was 9.91% WER on the test set, improving the WER of the raw ASR system with 5.66%.

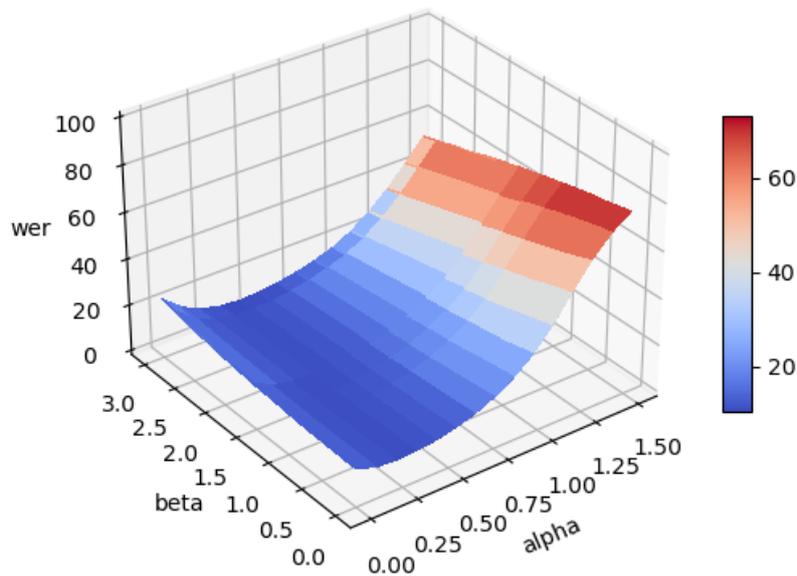

**Figure 2.** Surface plot of the WER with respect to the KenLM hyperparameters.

The other correction components were created mainly for the purposes of the ROBIN project, considering the envisaged interaction scenarios (short questions associated with closed worlds). Therefore, they were not currently evaluated on the general ASR test set and it is envisaged to be later evaluated on a dedicated ROBIN set. Nevertheless, the architecture presented in Section 3 is valid and any improvements can be realized at module level. Furthermore, in order to test the integration capabilities, the modules were integrated in the RELATE platform (Păiș et al., 2019), allowing users to upload a recorded wav file or make a recording directly in the platform and run it through the ASR system. The predicted transcription can then be analysed using the available annotation mechanisms within the RELATE platform. A picture of the Robin ASR in the RELATE platform is depicted in Figure 3.

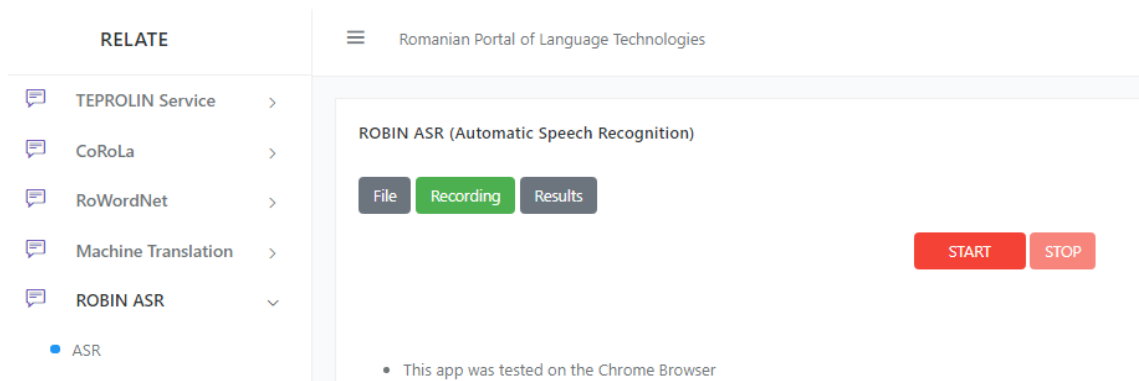

**Figure 3.** Picture of Robin ASR system in the RELATE platform.

## 8. Conclusions and Future Work

This paper presents the results of experimenting with an automatic speech recognition system for Romanian language, in the context of the ROBIN project. Our main goal was to obtain a low latency system, usable in near real-time applications such as voice interaction with a robot. However, in addition to the low latency (70 ms average recognition time) we managed to obtain a system with very good performance, 9,91% WER on our balanced test split.

The system was exposed as a REST API service, complemented by additional text correction modules, also available as individual services. The proposed integration scheme into more complex systems was validated by a first integration in the RELATE platform. The code is available at the following link: https://github.com/racai-ai/RobinASR together with the open-sourced release of the dialog manager (Ion et al., 2020): https://github.com/racai-ai/ROBINDialog .

As future work we consider integrating the Romanian BERT (Dumitrescu et al., 2020) in the postprocessing pipeline and use the model as a replacement for the n-gram language model to correct the transcription inferred by the ASR module, as presented in (Hrinchuk et al., 2019). However, the integration of BERT might add considerable latency overhead due to its high computational cost, so a pruning technique should be applied beforehand.

*Acknowledgements*

This work was realized in the context of the ROBIN project, a 33 months grant of the Ministry of Research and Innovation PCCDI-UEFISCDI, project code PN-III-P1-1.2-PCCDI-2017-734 within PNCDI III.

## References

Amodei, D., Ananthanarayanan, S., Anubhai, R., Bai, J., Battenberg, E., Case, C., Casper, J., Catanzaro, B., Cheng, Q., Chen, G. and Chen, J. (2016). June. Deep speech 2: End-to-end speech recognition in English and Mandarin. In *International conference on machine learning*, New York City, USA, 173-182.

Ardila, R., Branson, M., Davis, K., Henretty, M., Kohler, M., Meyer, J., Morais, R., Saunders, L., Tyers, F.M. and Weber, G. (2019). Common voice: A massivelymultilingual speech corpus. *arXiv:1912.06670*.

Avram, A.M., Păiș, V., Tufiș, D. (2020) Towards a Romanian end-to-end automatic speech recognition based on DeepSpeech2. In *Proceedings of the Romanian Academy*, Series A, in-print.

Bański, P., Fischer, P., Frick, E., Ketzan, E., Kupietz, M., Schnober, C., Schonefeld, O., Witt, A. (2012). The New IDS Corpus Analysis Platform: Challenges and Prospects. In *Proceedings of the 8th International Conference on Language Resources and Evaluation (LREC 2012)*, 2905-2911.

Brown, T.B., Mann, B., Ryder, N., Subbiah, M., Kaplan, J., Dhariwal, P., Neelakantan, A., Shyam, P., Sastry, G., Askell, A. and Agarwal, S. (2020). Language models are few-shot learners. *arXiv:2005.14165*.

Collobert, R., Puhrsch, C. and Synnaeve, G. (2016). Wav2letter: an end-to-end convnet-based speech recognition system. *arXiv:1609.03193*.

Cristea, D., Diewald, N., Haja, G., Mărănduc, C., Mititelu, V.B., Onofrei M. (2019). How to find a shining needle in the haystack. Querying CoRoLa: solutions and perspectives. In *Revue Roumaine de linguistique*, LXIV (3).

Cristea, D., Pistol, I., Boghiu, Ș., Bibiri, A.D., Gîfu, D., Scutelnicu, A., Onofrei, M., Trandabăț, D., Bugeag, G. (2020). CoBiLiRo: A Research Platform for Bimodal Corpora. In *Proceedings of the 1st International Workshop on Language Technology Platforms (IWLTP 2020)*, 22–27, Language Resources and Evaluation Conference (LREC 2020), Marseille, France.

Di Gangi, M.A., Negri, M. and Turchi, M. (2019). Adapting Transformer to end-to-end spoken language translation. In *Proceedings INTERSPEECH*, Graz, Austria, 1133-1137.

Dumitrescu, S.D., Avram, A.M. and Pyysalo, S., (2020). The birth of Romanian BERT. In *Proceedings of the 2020 Conference on Empirical Methods in Natural Language Processing: Findings*, Association for Computational Linguistics, 4324–4328.

Dumitrescu, S.D., Boroș, T. and Ion, R. (2014). Crowd-Sourced, Automatic Speech-Corpora Collection–Building the Romanian Anonymous Speech Corpus. In *CCURL 2014: Collaboration and Computing for Under-Resourced Languages in the Linked Open Data Era*, Reykjavik, Iceland, 90-94.

Garg, A. and Sharma, P. (2016). Survey on acoustic modeling and feature extraction for speech recognition. In *2016 3rd International Conference on Computing for Sustainable Global Development*, New Delhi, India, 2291-2295.

Georgescu, A., Cucu, H., Burileanu, C. (2019) Kaldi-based DNN Architectures for Speech Recognition in Romanian. In *Proceedings of the 2019 International Conference on Speech Technology and Human-Computer Dialogue (SpeD)*, Timisoara, Romania, pp. 1-6, doi: 10.1109/SPED.2019.8906555.

Georgescu, A.L., Caranica, A., Cucu, H. and Burileanu, C. (2018). RODIGITS-a Romanian connected-digits speech corpus for automatic speech and speaker recognition. *University Politehnica of Bucharest Scientific Bulletin,* Bucharest, Romania.

ROMANIAN SPEECH RECOGNITION EXPERIMENTS FROM THE ROBIN PROJECT

Georgescu, A.L., Cucu, H., Buzo, A. and Burileanu, C. (2020). RSC: A Romanian Read Speech Corpus for Automatic Speech Recognition. In *Proceedings of the 12th Language Resources and Evaluation Conference*, Marseille, France, 6606-6612.

Graves, A., Fernández, S., Gomez, F. and Schmidhuber, J. (2006). Connectionist temporal classification: labelling unsegmented sequence data with recurrent neural networks. In *Proceedings of the 23rd international conference on Machine learning*, Pittsburgh, United States, 369-376.

Graves, A., Fernández, S., Gomez, F. and Schmidhuber, J. (2006). Connectionist temporal classification: labelling unsegmented sequence data with recurrent neural networks. In *Proceedings of the 23rd international conference on Machine learning*, 369-376.

He, K., Zhang, X., Ren, S. and Sun, J. (2016). Deep residual learning for image recognition. In *Proceedings of the IEEE conference on computer vision and pattern recognition*, Las Vegas, USA, 770-778.

Heafield, K. (2011). KenLM: Faster and smaller language model queries. In *Proceedings of the sixth workshop on statistical machine translation*, Edinburgh, Scotland, 187-197.

Hochreiter, S. and Schmidhuber, J., (1997). Long short-term memory. In *Neural computation*, 1735-1780.

Hrinchuk, O., Popova, M. and Ginsburg, B. (2020). Correction of Automatic Speech Recognition with Transformer Sequence-To-Sequence Model. In *Proceedings of the IEEE International Conference on Acoustics, Speech and Signal Processing (ICASSP)*, 7074-7078.

Ioffe, S. and Szegedy, C. (2015). Batch Normalization: Accelerating Deep Network Training by Reducing Internal Covariate Shift. In *International Conference on Machine Learning*, 448-456.

Ion, R., Badea, V.G., Cioroiu, G., Barbu Mititelu, V., Irimia, E., Mitrofan, M., and Tufiș, D. (2020). A Dialog Manager for Micro-Worlds. In *Studies in informatics and control*, vol.29, issue 4 (in print).

Kingma, D.P. and Ba, J. (2015). Adam: A method for stochastic optimization. In *Proceedings of the 3rd International Conference on Learning Representations ICLR*.

Krizhevsky, A., Sutskever, I. and Hinton, G.E. (2012). Imagenet classification with deep convolutional neural networks. In *Advances in neural information processing systems*, Harrahs and Herveys, USA, 1097-1105.

LeCun, Y., Bottou, L., Bengio, Y. and Haffner, P. (1998). Gradient-based learning applied to document recognition. In *Proceedings of the IEEE*, 2278-2324.

Logan, B. (2000). Mel frequency cepstral coefficients for music modeling. In *Proceedings of the International Symposium on Music Information Retrieval ISMIR*, 1-11.

Lopatovska, I., Rink, K., Knight, I., Raines, K., Cosenza, K., Williams, H., Sorsche, P., Hirsch, D., Li, Q. and Martinez, A. (2019). Talk to me: Exploring user interactions with the Amazon Alexa. In *Journal of Librarianship and Information Science*, United Kingdom, 984-997.

Mititelu, V.B., Tufiș, D. and Irimia, E. (2018). The reference corpus of the contemporary romanian language (CoRoLa). In *Proceedings of the Eleventh International Conference on Language Resources and Evaluation*, Miyazaki, Japan.

Noda, K., Yamaguchi, Y., Nakadai, K., Okuno, H.G. and Ogata, T. (2015). Audio-visual speech recognition using deep learning. In *Applied Intelligence*, 722-737.

Păiș, V., Tufiș, D., Ion, R. (2019) Integration of Romanian NLP tools into the RELATE platform. In *Proceedings of the International Conference on Linguistic Resources and Tools for Processing Romanian Language – CONSILR 2019*, pages 181-192.

Park, D.S., Chan, W., Zhang, Y., Chiu, C.C., Zoph, B., Cubuk, E.D. and Le, Q.V. (2019). Specaugment: A simple data augmentation method for automatic speech recognition. *arXiv:1904.08779*.

Pascanu, R., Mikolov, T. and Bengio, Y., (2013). On the difficulty of training recurrent neural networks. In *International conference on machine learning*, Atlanta, United States, 1310-1318.

Russakovsky, O., Deng, J., Su, H., Krause, J., Satheesh, S., Ma, S., Huang, Z., Karpathy, A., Khosla, A., Bernstein, M., Berg, A.C., Fei-Fei, L. (2015). ImageNet Large Scale Visual Recognition Challenge. In *International Journal of Computer Vision (IJCV)*, vol. 115, no. 3, pp. 211-252.

Stan, A., Dinescu, F., Țiple, C., Meza, Ș., Orza, B., Chirilă, M. and Giurgiu, M. (2017). The SWARA speech corpus: A large parallel Romanian read speech dataset. In *2017 International Conference on Speech Technology and Human-Computer Dialogue (SpeD)*, Bucharest, Romania, 1-6.

Stan, A., Yamagishi, J., King, S. and Aylett, M. (2011). The Romanian speech synthesis (RSS) corpus: Building a high quality HMM-based speech synthesis system using a high sampling rate. In *Speech Communication*, 442-450.

Suárez, P.J.O., Sagot, B. and Romary, L. (2019). Asynchronous pipeline for processing huge corpora on medium to low resource infrastructures. In *7th Workshop on the Challenges in the Management of Large Corpora*.

Tufiș, D., Mititelu, V.B., Irimia, E., Păiș, V., Ion, R., Diewald, N., Mitrofan, M., Onofrei, M. (2019b). Little strokes fell great oaks. Creating CoRoLa, the reference corpus of contemporary Romanian. In *Revue Roumaine de linguistique*, LXIV (3).

Tufiș, D., Barbu Mititelu, V., Irimia, E., Mitrofan, M., Ion, R. and Cioroiu, G.(2019a). Making Pepper Understand and Respond in Romanian. In *Proceedings of the 22nd International Conference on Control Systems and Computer Science*, 682-688.

Xiang, X., Qian, Y. and Yu, K., (2017). Binary Deep Neural Networks for Speech Recognition. In *Proceedings INTERSPEECH*, 533-537.

You, K., Long, M., Wang, J. and Jordan, M.I. (2019). How Does Learning Rate Decay Help Modern Neural Networks?. *arXiv:1908.01878*.